\pdfoutput=1
\AddToHook{cmd/maketitle/after}{%
  \begingroup
  \footnotetext{\raggedright Findings of the Association for Computational \mbox{Linguistics: EACL 2027.}}%
  \endgroup
}

\documentclass[11pt]{article}

\usepackage[final]{acl}

\usepackage[T1]{fontenc}
\usepackage[utf8]{inputenc}
\usepackage{times}
\usepackage{latexsym}
\usepackage{amsmath,amssymb,amsthm}
\usepackage{graphicx}
\usepackage{booktabs}
\usepackage{multirow}
\usepackage{xcolor}
\usepackage[normalem]{ulem}
\usepackage{microtype}
\usepackage[capitalize,noabbrev]{cleveref}

\def\FinalCopy{}

\theoremstyle{plain}

\newtheorem{lemma}{Lemma}
\newtheorem{corollary}{Corollary}
\newtheorem{proposition}{Proposition}

\theoremstyle{definition}

\theoremstyle{remark}

\crefname{section}{Section}{Sections}
\crefname{subsection}{Section}{Sections}
\crefname{figure}{Figure}{Figures}
\crefname{table}{Table}{Tables}
\crefname{theorem}{Theorem}{Theorems}
\crefname{lemma}{Lemma}{Lemmas}
\crefname{corollary}{Corollary}{Corollaries}
\crefname{equation}{Eq.}{Eqs.}
\crefname{appendix}{Appendix}{Appendices}

\newcommand{\appref}[1]{Appendix~\ref{#1}}

\ifdefined\FinalCopy
  \DeclareRobustCommand{\revision}[1]{#1}
  \newenvironment{revised}{}{}
\else
  \definecolor{revisionblue}{HTML}{145AB6}
  \DeclareRobustCommand{\revision}[1]{\textcolor{revisionblue}{#1}}
  \newenvironment{revised}{\color{revisionblue}}{}
\fi
\title{Predicting Inference-Time Scaling Gains from Labeled Validation-Set Output Statistics}

\author{
  Luyang Zhang \\
  Carnegie Mellon University \\
  \texttt{luyangz@andrew.cmu.edu} \\
  \And
  Jingyan Li \\
  Johns Hopkins University \\
  \texttt{jli336@alumni.jh.edu} \\
}

\begin{document}
\raggedbottom

\maketitle

\begin{abstract}
\revision{Verifier-selected best-of-$N$ scaling} (drawing $N$ candidate answers from a language model and returning the one a reward model ranks highest) improves accuracy by an amount that varies across models, but \revision{measuring that gain directly requires both candidate generation and reward-model scoring.}
Prior work links cheap statistics of a model's sampled outputs and validation-set correctness (how often samples agree, how diverse they are, how confident the model is, and where correct samples appear) to model behavior, but does not isolate which of these form a \revision{compact predictor} of best-of-$N$ gain.
We fit ridge predictors on \revision{features from one round of sampling on a labeled validation set}, use bootstrap-Lasso \revision{to measure how often candidate features are selected across resamples}, and give \revision{an error bound that accounts for the limits of a linear predictor}.
\revision{Using only four features, our predictor achieves a Spearman rank correlation of $0.90$ with measured best-of-$N$ gains. The evaluation spans math and code configurations from three base-model families and six post-training methods.}
The intended use is labeled validation-set screening of candidate configurations \revision{after the shared generation pass and} before paying the full reward-model scoring cost.
\end{abstract}

\section{Introduction}
\label{sec:introduction}

Inference-time scaling (drawing many candidate answers from a language model and selecting one with a verifier or by majority vote) \revision{is widely used when deploying} large language models (LLMs) on reasoning-heavy tasks. However, \revision{its gains vary} across models and tasks \citep{Cobbe2021Training, Wang2022Self, Snell2024Scaling}, and \revision{it remains difficult to predict} whether scaling will help on a new model-task pair. Because inference-time scaling \revision{is computationally expensive}, running it without such a predictor \revision{can waste compute when the gain is small}. This raises \revision{the question of which output statistics from a labeled validation set} can predict, \revision{at low cost}, whether \revision{verifier-selected best-of-$N$} will improve accuracy.

\revision{Two lines of work address parts of} this question. The first measures inference-time scaling gain directly across models and tasks by running scaling end-to-end \citep{Snell2024Scaling, Brown2024Large, Wu2024Inference}, \revision{showing how gains vary} but offering no efficient predictor. The second \revision{computes inexpensive statistics} from a model's sampled outputs, such as agreement, diversity, and confidence \citep{Kadavath2022Language, Holtzman2020Nucleus, Wang2022Self}, and links them to model behavior but not to scaling gain. What is missing is \revision{a predictor of scaling gain that uses statistics from a labeled validation set, generalizes} across base models and post-training methods, comes with an explicit error decomposition, and identifies \revision{which statistics help predict the gain when used together}.

\begin{figure*}[t]
  \centering
  \includegraphics[width=\textwidth]{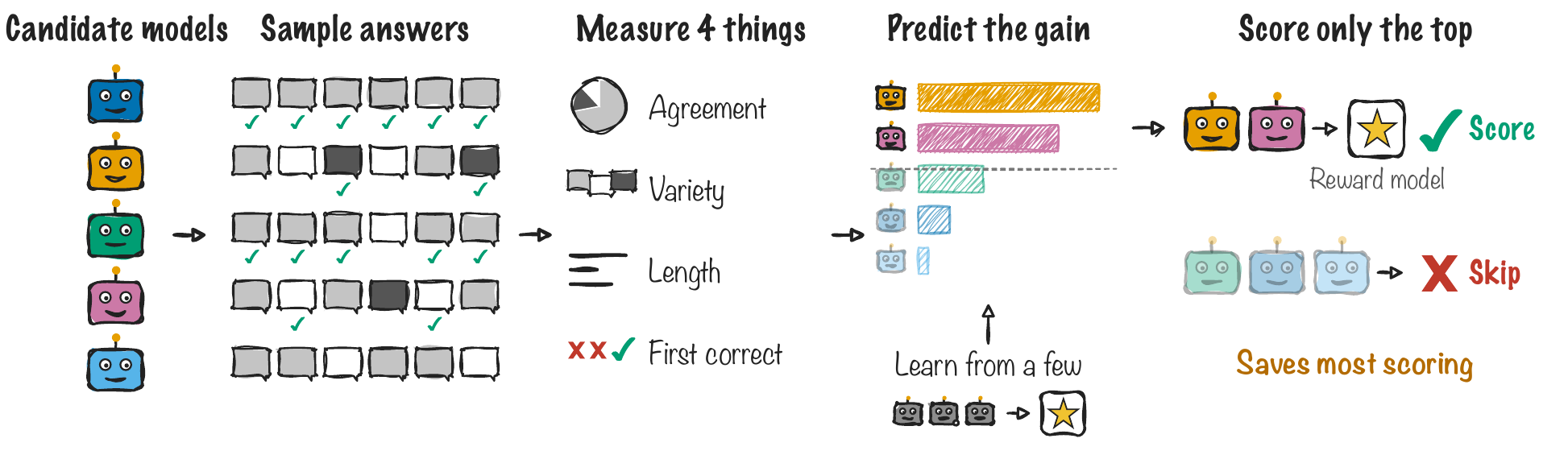}
  \caption{\revision{We use four features from sampled validation answers to predict best-of-$N$ gains. We fit the predictor on a small set of configurations with measured gains, then use the reward model to score the top-ranked configurations.}}
  \label{fig:overview}
\end{figure*}

Our framework links low-cost validation-set sample statistics directly to scaling-gain prediction \revision{(\cref{fig:overview})}. For each configuration of base model, RL method, task domain, and seed, we sample from the model at three temperatures on labeled held-out prompts and compute statistics in two groups: label-free summaries of how often the model produces the same answer across samples, and validation-assisted summaries of how agreement and correctness vary from prompt to prompt. These features describe a model's behavior at \revision{low additional cost once the answers have been generated. We run the reward model to measure gains for the configurations used to train the predictor and for those selected for full evaluation.}

We then fit a single ridge regression jointly over all three temperatures and use bootstrap-Lasso \revision{to see which features are repeatedly selected when we fit to verifier-selected gains.} A concentration analysis decomposes the regression's error into an explicit linear-approximation residual, feature-side uncertainty in the sampled statistics, and target-side uncertainty when the population gain is replaced by its empirical estimate; \revision{the bound also accounts for error in the fitted coefficients.}

\revision{First, the compact predictor ranks configurations by how much they benefit from best-of-$N$. On the evaluated math and code configurations, its predictions have a Spearman correlation of $0.90$ with measured gains. It ranks gains more accurately than agreement rate alone.}

\revision{Second, ranking by predicted gain helps select the configurations that benefit most. In a separate experiment with the 18-feature ridge, selecting the top five gives a mean precision of $0.90$.}

\revision{Third, the predictor can rank gains when the post-training method or reward model changes. Holding out one post-training method at a time gives Spearman correlations from $0.78$ to $0.94$. A predictor fitted to gains measured by one reward model reaches $\rho=0.85$ on gains measured by a second without refitting.}

We make three contributions.
\begin{itemize}
  \item \revision{\textbf{Gain prediction.} We predict best-of-$N$ gains from statistics of sampled answers on a labeled validation set. We examine which features recur across resamples and derive an error bound for the predictor.}
  \item \revision{\textbf{Feature efficiency.} We show that four validation-set features are enough to predict which configurations benefit most from verifier-selected best-of-$N$ scaling.}
  \item \revision{\textbf{Scope and failure modes.} We find that the predictor ranks verifier-selected gains well on the combined math and code evaluation set. In the earlier 18-feature analysis, rank correlation is near zero for majority-vote gains and negative when we hold out the entire programming domain.}
\end{itemize}

\revision{The screening code is available on \href{https://github.com/LZhang12345/bon-gain-screening}{GitHub}.}

\section{Related Work}
\label{sec:related-work}

\textbf{Inference-time scaling.}
Best-of-$N$ with a verifier was introduced for grade-school math by \citet{Cobbe2021Training}, and self-consistency, which selects the majority answer among sampled chains of thought, was popularized by \citet{Wang2022Self}.
Stronger verifiers widen the gain further, with process reward models trained on step-level annotations \citep{Lightman2023Lets, Uesato2022Solving} sometimes adding tens of points over a single sample.
\citet{Snell2024Scaling} study when test-time compute helps and show that the optimal allocation depends on prompt difficulty and base-model competence.
\revision{These works improve best-of-$N$ or decide how much to sample for a given model, whereas we predict how much each configuration will gain from best-of-$N$ before any reward-model scoring.}

\revision{\textbf{Predicting behavior from low-cost measurements.}}
Scaling laws relate loss to parameters, data, and compute \citep{Kaplan2020Scaling, Hoffmann2022Training}. Downstream accuracy can decrease with model scale on some tasks \citep{McKenzie2023Inverse}, and the choice of metric can create or hide apparent ability jumps \citep{Schaeffer2023Are}.
A closer line predicts capabilities from properties of the trained model itself.
\citet{Burnell2023Revealing} factor benchmark results into latent skills, and \citet{Ruan2024Observational} fit observational scaling laws across checkpoints to extrapolate task accuracy.
Two studies that examine heterogeneity in BoN benefit, \citet{Brown2024Large} and \citet{Wu2024Inference}, characterize the gain after running BoN on each model.
\revision{In interpretability, cheap gradient estimates of intervention effects can be wrong \citep{zhang2026attributionpatchingliesdiagnosis}, and a few exact measurements can correct them \citep{zhang2026correctingattributionpatchingexact}.}
\revision{These works forecast accuracy or measure the gain after running best-of-$N$, whereas we predict the gain beforehand from a model's sampled answers.}

\textbf{Output distribution and probing.}
Calibration work shows that an LLM's confidence and entropy carry information about correctness \citep{Kadavath2022Language, Jiang2021How}.
Probing recovers task-relevant variables that are not obvious from outputs, including unsupervised elicitation of truthfulness \citep{Burns2023Discovering} and internal activations that track whether assertions are correct \citep{Azaria2023Internal}.
\revision{Token probabilities can also estimate how user preferences are distributed over categories \citep{zhang2026spectrarevealingspectrumuser}.}
At the output level, sample-diversity measures such as self-BLEU \citep{Zhu2018Texygen} have long been used to characterize generative models, and self-consistency \citep{Wang2022Self} is itself a one-feature summary of agreement across stochastic samples.
RLHF can reduce output diversity \citep{Kirk2024Understanding}, which can affect the benefit of drawing more candidates.
\revision{Post-training can also change model behavior within a base-model family. For example, \citet{zhang2026posttrainingrecipemodelfamily} find differences in multi-agent dialogue across post-training methods.}
\revision{These works use such signals to judge single answers or describe model behavior, whereas we aggregate them to predict best-of-$N$ gain.}

\section{Framework and Theoretical Analysis}
\label{sec:method}

This section defines configurations and sampled-output statistics (\cref{sec:method:problem}), describes \revision{the predictor and how it is used for screening} (\cref{sec:method:predictor}), and gives \revision{an error bound and its implications for ranking} (\cref{sec:method:analysis}).

\subsection{Configuration, Gain, and Features}
\label{sec:method:problem}

\revision{We define a configuration by its trained model and training conditions. Our primary measure of gain is the difference between best-of-$N$ accuracy and single-sample accuracy. To predict that gain, we compute statistics of sampled answers on a labeled validation set.}

\revision{Prior work shows that scaling gains vary with the model and task \citep{Brown2024Large, Wu2024Inference, Snell2024Scaling}. We also vary the post-training method. RLHF and best-of-$N$ can also change output diversity and generalization \citep{Kirk2024Understanding}. We write a configuration as $c=(\pi_\theta,\mathrm{PT},\mathcal{D},s)$, where $\pi_\theta$ is the fine-tuned model, $\mathrm{PT}$ is the post-training method, $\mathcal{D}$ is the task domain, and $s$ is the training seed when multiple runs are available. Let $k$ be the number of candidates sampled per prompt. We use $\mathrm{BoN}@k$ for the accuracy of the candidate ranked highest by the reward model, so $N=k$ in best-of-$N$. Majority voting returns the most frequent answer and has accuracy $\mathrm{MV}@k$. We estimate single-sample accuracy, $\mathrm{pass}@1$, as mean correctness across the same $k$ candidates \citep{Chen2021Evaluating}.}

\textbf{How we summarize the gain.} We measure best-of-$N$ improvement as the difference in accuracy $G_{\mathrm{add}} \equiv \mathrm{BoN}@k - \mathrm{pass}@1$ \citep{Cobbe2021Training, Brown2024Large}. We use it as the primary target because it counts extra correct answers per prompt, and also consider three reparameterizations,
\begin{itemize}
  \setlength{\itemsep}{1pt}
  \setlength{\parsep}{0pt}
  \setlength{\topsep}{2pt}
  \setlength{\partopsep}{0pt}
  \setlength{\leftmargin}{1.4em}
  \item $G_{\mathrm{mult}} \equiv \mathrm{BoN}@k / \mathrm{pass}@1$ (multiplicative),
  \item $G_{\mathrm{norm}} \equiv (\mathrm{BoN}@k - \mathrm{pass}@1)/(1 - \mathrm{pass}@1 + \varepsilon_0)$ (fraction of remaining gap closed; the small constant $\varepsilon_0 > 0$ \revision{keeps the denominator positive when single-sample accuracy approaches one}),
  \item $G_{\mathrm{log}} \equiv \log\mathrm{BoN}@k - \log\mathrm{pass}@1$ (log-ratio),
\end{itemize}
plus the majority-voting variant $G_{\mathrm{MV}} \equiv \mathrm{MV}@k - \mathrm{pass}@1$.
\revision{Each gain function $G(a,b)$ compares a selection accuracy $a$ with single-sample accuracy $b$. For the four best-of-$N$ gain measures above, we pick the answer with the highest reward-model score and use the resulting accuracy, $a=\mathrm{BoN}@k$. For the majority-vote measure, we pick the most frequent answer and use $a=\mathrm{MV}@k$. The bound below applies to the family of gains that are Lipschitz on the full accuracy range.}
\begin{multline}
\label{eq:gain-class}
\mathcal{G}_L = \bigl\{ G : G \text{ is } L_G\text{-Lipschitz on } [0,1]^2 \\
\text{for some finite } L_G \bigr\}.
\end{multline}
The additive gain
\begin{equation}
\label{eq:gain}
g(c, T) \;=\; \mathrm{BoN}@k(c, T) \;-\; \mathrm{pass}@1(c, T)
\end{equation}
is Lipschitz on $[0,1]^2$ with constant $L_g = \sqrt{2}$.

\revision{\textbf{Output-distribution features.} We first compare the extracted answers for each prompt. Agreement rate is the fraction of sample pairs that give the same answer. Majority fraction is the share of samples that give the most frequent answer. If three of four samples give one answer and the fourth gives another, three of the six pairs match, so agreement rate is $1/2$ and majority fraction is $3/4$. We average both statistics across prompts. We also average answer entropy across prompts. Entropy is higher when samples are spread more evenly across different answers.}

\revision{We compare the full completions as well. Our self-BLEU proxy measures the share of distinct four-word sequences that two completions have in common \citep{Zhu2018Texygen}. The unique-bigram ratio divides the number of distinct adjacent word pairs by the total number of such pairs across completions for a prompt. A lower ratio indicates more repetition. Embedding similarity averages the cosine similarity between the completions' vector representations. The log-probability features summarize how much probability the model assigns to its generated tokens and how those scores vary within and across completions \citep{Kadavath2022Language}. We also record how much probability the model puts on its most likely next tokens. \Cref{tab:feature-catalog} lists all ten features.}

\revision{\textbf{Additional sample statistics.} Averages over prompts can hide how differently a model behaves on easy and hard prompts, so we add statistics that capture this variation. For example, majority-fraction spread is the standard deviation, across prompts, of the fraction of samples that give the most frequent answer. A large spread means the model is consistent on some prompts and scattered on others. This group also includes the spread of completion lengths, the median per-prompt answer entropy, and the median position of the first correct sample, the only feature that needs gold answers (\cref{tab:feature-catalog}). \Cref{sec:experiments:setup} lists the four features used by the compact predictor.}

\subsection{\revision{Predictor and Screening}}
\label{sec:method:predictor}

\revision{The predictor maps a configuration's features to its gain. Both change with the sampling temperature, so one observation is a configuration at one temperature, with its features and gain computed over the labeled validation prompts. We pool observations from several temperatures in one ridge regression,}
\begin{equation}
\label{eq:joint}
g(c, T) \;=\; \boldsymbol{\beta}^\top \mathbf{x}(c, T) \;+\; \gamma\,(T - T_0) \;+\; \varepsilon(c, T),
\end{equation}
\revision{where $\mathbf{x}(c,T)\in\mathbb{R}^d$ holds the $d$ chosen features, $\boldsymbol{\beta}$ holds their coefficients, $\gamma$ is a temperature coefficient, $T_0$ is a reference temperature, and $\varepsilon(c,T)$ is the residual. Sharing $\boldsymbol{\beta}$ across temperatures adds observations without adding parameters.}

\revision{\textbf{Screening.} The regression is fit on a small support set of configurations whose gains have been measured with the verifier. For each new configuration, we generate candidates, compute the features, and predict the gain. Only the configurations predicted to gain most are then scored with the verifier.}

\subsection{Theoretical Analysis}
\label{sec:method:analysis}

\revision{The bound separates three sources of prediction error. First, even the best linear prediction can differ from the true gain. Second, we estimate the features from a finite set of prompts and sampled answers. Third, we fit the coefficients using a finite set of training configurations. Comparing a prediction with the measured gain adds a fourth source of error. We compute that gain from estimates of best-of-$N$ accuracy and single-sample accuracy. For a Lipschitz gain function, $L_G$ bounds how much the gain can change for a given error in those two estimates.}

\revision{\textbf{Setup.} For a configuration $c$ at temperature $T$, the population gain $G(c,T)$ is computed from the population best-of-$N$ and single-sample accuracies, $\mathrm{BoN}@k(c,T)$ and $\mathrm{pass}@1(c,T)$. We estimate both accuracies and the gain from $P$ prompts and $n_{\text{samp}}$ completions per prompt. To analyze the full predictor in \cref{eq:joint}, we include $T-T_0$ in $\mathbf{x}(c,T)$ and $\gamma$ in $\boldsymbol{\beta}$. Here $d$ counts the temperature coordinate. We write $\hat{\mathbf{x}}$ for the estimated vector. We write $\hat{\boldsymbol\beta}\in\mathbb{R}^d$ for the ridge coefficients fitted on $m$ configurations and $\boldsymbol\beta^\star$ for the best linear coefficients in the population.}

\revision{\textbf{Sampling assumptions.} For each configuration and temperature, we treat the prompts as independent draws from the same task distribution. Given a prompt, we assume the completions are drawn independently from the same sampling distribution. Each prompt contributes an accuracy value in $[0,1]$. Independence and boundedness are the standard conditions for applying Hoeffding's inequality to these averages \citep{Hoeffding1963Probability}. Configurations can share prompts because we combine their probability bounds with a union bound.}

\revision{\textbf{Feature assumptions.} The bound uses sampled feature values in $[0,1]$, with fixed rescaling where needed. We use a separate confidence bound $\varepsilon_j$ for the estimation error of each sampled feature $j$ (\appref{app:proof}). Temperature is known, so its coordinate has zero estimation error. The coefficient-weighted feature error is $E_\beta(\hat{\boldsymbol{\beta}}):=\sum_j|\hat\beta_j|\varepsilon_j$. We write $\sigma_{\text{tgt}}$ for the Hoeffding bound on the joint error of the two accuracy estimates. We define the linear approximation error as}
\[
A_G^\star := \sup_{c,T}\bigl|\boldsymbol{\beta}^{\star\top}\mathbf{x}(c,T) - G(c,T)\bigr|
\]
over the training configurations and temperatures.

\begin{proposition}[Joint concentration on training configurations, Lipschitz family]
\label{thm:joint-bound}
Let $G \in \mathcal{G}_L$ be a Lipschitz gain function with constant $L_G$, and fix $\delta \in (0, 1)$ split symmetrically as $\delta_{\text{tgt}} = \delta_{\text{feat}} = \delta/2$. \revision{Under the sampling and feature assumptions above}, with probability at least $1-\delta$ simultaneously over all $m$ training configurations $c$ and temperatures $T$,
\begin{align}
\label{eq:joint-bound}
\bigl|\,\hat{\boldsymbol{\beta}}^{\!\top}\hat{\mathbf{x}}(c, T) &- G(c, T)\,\bigr| \;\leq\; \underbrace{A_G^\star}_{\text{approx.}} \;+\; \underbrace{E_\beta(\hat{\boldsymbol{\beta}})}_{\text{\revision{features}}} \notag \\
&\;+\; \underbrace{\|\hat{\boldsymbol{\beta}} - \boldsymbol{\beta}^\star\|_2 \cdot \|\mathbf{x}(c,T)\|_2}_{\text{coefficient}}.
\end{align}
\revision{The error in the measured gain satisfies $|\hat G(c,T)-G(c,T)|\leq L_G\,\sigma_{\text{tgt}}$. We add this term to the right-hand side when comparing predictions to measured gains. The primary gain $g=G_{\mathrm{add}}$ has $L_g=\sqrt{2}$. The bound on the measured gain requires $G\in\mathcal{G}_L$. The multiplicative and log-ratio gains are non-Lipschitz near $\mathrm{pass}@1=0$. We test prediction for those gains in \cref{sec:experiments:robustness}.}
\end{proposition}

The full proof is given in \appref{app:proof}.

\begin{corollary}[Conditional held-out transfer]
\label{cor:heldout-transfer}
Assume the training and held-out configurations are drawn i.i.d.\ from a common population, $\|\mathbf{x}(c,T)\|_2 \leq R_x$, and the same feature-concentration radii $\varepsilon_j$ hold for an independent held-out configuration. Suppose further that the population approximation residual is bounded on that support,
\[
A_{G,\mathrm{pop}} := \sup_{c,T}\bigl|\boldsymbol{\beta}^{\star\top}\mathbf{x}(c,T)-G(c,T)\bigr|,
\]
and write $\eta_m := C_{\mathrm{ridge}}\sqrt{(d+\log(1/\delta_{\mathrm{est}}))/m}$ \revision{and assume $\|\hat{\boldsymbol{\beta}}-\boldsymbol{\beta}^\star\|_2 \leq \eta_m$ holds with probability at least $1-\delta_{\mathrm{est}}$.} Then, for a fresh held-out configuration at a fixed $T$, with probability at least $1-\delta-\delta_{\mathrm{est}}$,
\begin{align}
\bigl|\,\hat{\boldsymbol{\beta}}^{\!\top}\hat{\mathbf{x}}(c,T) - \hat G(c,T)\,\bigr|
&\leq A_{G,\mathrm{pop}} + E_\beta(\hat{\boldsymbol{\beta}}) \notag \\
&\quad + R_x\eta_m + L_G\sigma_{\mathrm{tgt}} .
\end{align}
\end{corollary}

\revision{The held-out bound assumes a fresh configuration from the same population as the training configurations. It also assumes the coefficient error bound $\eta_m$. Random-design ridge analyses use additional conditions on the features and the noise in the response to derive their bounds \citep{Hsu2012RandomDesign}. The assumed coefficient error bound $\eta_m$ grows with the number of predictor inputs $d$ and shrinks as the number of training configurations $m$ increases. The experiments measure ranking accuracy and top-$K$ selection.}

\revision{Spearman correlation measures whether the predictor puts configurations in the right order for screening. The following lemma relates the error in individual scores to errors in that ranking.}

\begin{lemma}[Rank perturbation]
\label{lem:rank-perturb}
Let $m'\geq 2$ be the number of configurations and let $\hat{\boldsymbol r}$ and $\boldsymbol r^\star$ be their predicted and population gains. \revision{Assume $\|\hat{\boldsymbol r}-\boldsymbol r^\star\|_\infty \leq \Delta$ and break ties by a fixed ordering. Let $q_{\leq 2\Delta}$ be the fraction of pairs whose population gains differ by at most $2\Delta$. The Spearman correlation $\hat\rho$ between the resulting rankings satisfies $1-\hat\rho\leq 6q_{\leq 2\Delta}$.}
\end{lemma}

\revision{\textbf{Implication.} The lemma requires a bound on score error across the full set being ranked (\appref{app:bound-eval}). If the population gain gap between the $K$th and $(K+1)$st configurations exceeds $2\Delta$, the predicted top-$K$ set equals the population top-$K$ set.}

\section{Experiments}
\label{sec:experiments}

\revision{We first test whether predicted gains recover the ranking of configurations (\cref{sec:experiments:main}). We then ask whether the top-ranked configurations benefit more than a random selection and how much scoring the predictor requires (\cref{sec:experiments:utility}). We examine which features are repeatedly selected and test the effect of removing one entropy feature (\cref{sec:experiments:stability}). Finally, we vary the evaluation conditions and examine where the predictor makes larger errors (\cref{sec:experiments:robustness,sec:experiments:failures}).}

\subsection{Setup}
\label{sec:experiments:setup}

\revision{\textbf{Configurations.} We evaluate three base-model families at 2B--9B scale (Qwen2.5-3B/7B, Llama-3.2-3B and Llama-3.1-8B, and gemma-2-2B/9B). The post-training methods are DPO \citep{Rafailov2023DPO}, SimPO \citep{Meng2024SimPO}, KTO \citep{Ethayarajh2024KTO}, ORPO \citep{Hong2024ORPO}, GRPO \citep{Shao2024DeepSeekMath}, and supervised fine-tuning (SFT) without preference data \citep{Ouyang2022Training}. The main evaluation covers 40 math and 16 code configurations. A separate test holds out the entire code domain (\cref{sec:experiments:failures}). The sample sizes for each comparison are in \appref{app:setups}.}

\textbf{Sampling.} Each configuration generates $k = 64$ completions per prompt at three temperatures $T \in \{0.3, 0.7, 1.0\}$ on $P = 200$ labeled held-out prompts per domain (prompts that the post-trained model never saw during training). Full sampling and inference hyperparameters are in \appref{app:reproducibility}.

\revision{\textbf{Predictor.} We fit \cref{eq:joint} on the three temperatures with $T_0=0.7$ and choose the ridge penalty by inner cross-validation. The multi-feature baselines use the same regression, and the compact predictor uses the same four features in every fold. Two describe how the model's answers are distributed: majority-fraction spread and median per-prompt answer entropy. The other two measure variation in completion length and how early the model produces a correct answer. Only the first-correct-sample feature requires gold answers. See \appref{app:features} for the exact calculations.}

\revision{\textbf{Evaluation.} Our main test is leave-one-set-out (LOSO) evaluation, which holds out, one at a time, each group of configurations that share a base-model checkpoint and task domain. We use the remaining configurations to standardize features, select the ridge penalty, and fit the coefficients. We also report leave-one-configuration-out (LOO) evaluation, which holds out all temperatures of one configuration, and repeat LOSO with inner folds that keep each group together when selecting the penalty. For the intervals in the main comparison, the LOSO bootstrap resamples checkpoint and domain groups, and the LOO bootstrap resamples individual configurations. Both keep all temperatures of a configuration together.}

\revision{\textbf{Baselines.} We compare three types of baseline. Single-feature predictors use $\mathrm{pass}@1$ as an indicator of how much room there is to improve \citep{Snell2024Scaling}, mean reward-model score, mean per-token log-probability, sample diversity (self-BLEU \citealp{Zhu2018Texygen}), or first-token entropy. The output-distribution ridge combines ten features that measure answer concentration, diversity, and confidence \citep{Holtzman2020Nucleus, Zhu2018Texygen}. We also evaluate agreement rate alone and a version of the output-distribution ridge with the additional sample statistics. For top-$K$ selection, we compare with a uniformly random set of $K$ configurations and the oracle ranking by measured gain.}

\textbf{Target.} The best-of-$N$ scaling gain $g(c, T) = \mathrm{BoN}@k - \mathrm{pass}@1$ defined in \cref{eq:gain}, with Skywork-Reward-Llama-3.1-8B \citep{Liu2024Skywork} as the reward model that scores each sample and selects the highest-scoring one.

\revision{\textbf{Compute.} Screening and full evaluation start with the same generated answers. Screening computes features and predicts the gain on a CPU. Full evaluation instead scores each answer with Skywork, taking about 30 L40S GPU-minutes per configuration (\appref{app:reproducibility}).}

\subsection{Rank prediction}
\label{sec:experiments:main}

\begin{figure}[t]
  \centering
  \includegraphics[width=\columnwidth]{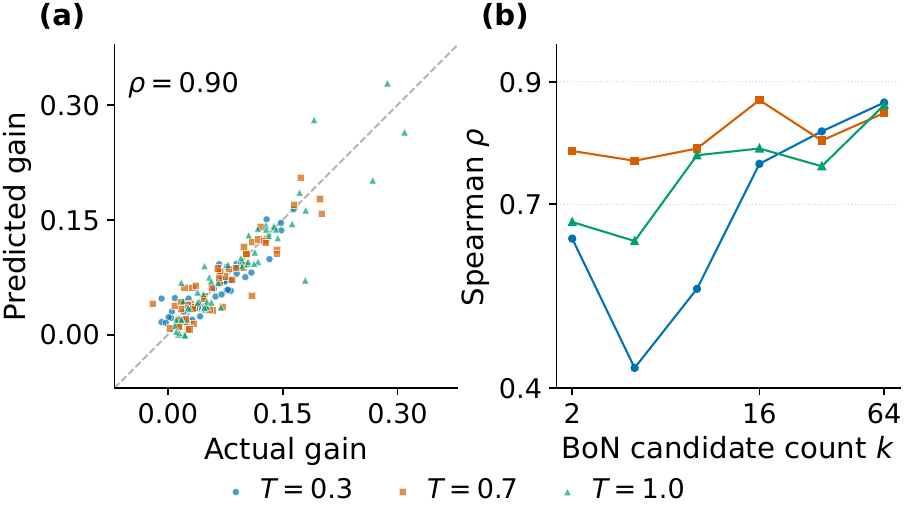}
  \caption{\revision{(a) Held-out predictions and measured gains for the compact predictor on 56 configurations, each evaluated at three temperatures. The diagonal marks equality. (b) LOSO Spearman $\rho$ for the 18-feature ridge on 32 configurations by candidate count $k$ and temperature. The inputs and $\mathrm{pass}@1$ use all 64 samples.}}
  \label{fig:calksweep}
\end{figure}

\revision{Across the held-out folds, the compact predictor ranks 56 configurations by measured gain with Spearman $\rho=0.90$ (\cref{tab:headline}). Keeping each checkpoint and domain group together in the inner folds gives $\rho=0.898$ on the same configurations. \Cref{fig:calksweep}a compares the compact predictor's held-out predictions with measured gains. The $y=x$ line shows where prediction and measurement agree.}

\begin{table}[h]
  \centering
  \caption{\revision{LOO and LOSO Spearman with 95\% cluster-bootstrap half-widths. The compact predictor uses 56 configurations. The multi-feature baselines use 50 configurations. The single-feature baselines use available values from that set, including 32 configurations for first-token entropy (\appref{app:setups}).}}
  \label{tab:headline}
  \small
  \setlength{\tabcolsep}{4pt}
  \renewcommand{\arraystretch}{1.25}
  \begin{tabular}{lcc}
    \toprule
    Feature set & LOO $\rho$ & LOSO $\rho$ \\
    \midrule
    \multicolumn{3}{l}{\textit{Multi-feature predictors}} \\
    \revision{Output-distribution ridge}                       & $0.89_{\pm .08}$ & $0.83_{\pm .21}$ \\[2pt]
    \quad \revision{$+$ Additional statistics}          & $\mathbf{0.89_{\pm .09}}$ & $\mathbf{0.87_{\pm .15}}$ \\[2pt]
    \revision{Compact predictor}            & $\mathbf{0.90_{\pm .05}}$ & $\mathbf{0.90_{\pm .13}}$ \\
    \midrule
    \multicolumn{3}{l}{\textit{Naive single-feature baselines}} \\
    $\mathrm{pass}@1$ alone                                   & $+0.12_{\pm .20}$ & $\!-0.24_{\pm .36}$ \\[4pt]
    \multirow{2}{*}{Self-BLEU}                                & \multirow{2}{*}{$+0.50_{\pm .15}$} & \multirow{2}{*}{$+0.31_{\pm .34}$} \\[3pt]
    \quad\quad\citep{Zhu2018Texygen}                            & & \\[4pt]
    First-token entropy                                      & $+0.28_{\pm .28}$ & $+0.14_{\pm .42}$ \\[4pt]
    Mean log-probability                                      & $+0.63_{\pm .12}$ & $+0.56_{\pm .23}$ \\[2pt]
    Mean reward-model score                                   & $+0.63_{\pm .15}$ & $+0.57_{\pm .33}$ \\[2pt]
    Agreement-rate feature                                    & $+0.62_{\pm .19}$ & $+0.57_{\pm .37}$ \\
    \bottomrule
  \end{tabular}
\end{table}

\textbf{Main correlation.} \revision{The output-distribution ridge reaches $\rho=0.83$, and adding the additional sample statistics raises it to $0.87$ on the same 50 configurations. The paired confidence interval for the difference between the compact predictor and the output-distribution ridge includes zero on that same set (\appref{app:bootstrap}).}

\revision{\textbf{Single-feature baselines.} In \cref{tab:headline}, single-feature predictors range from $\rho=-0.24$ for $\mathrm{pass}@1$ to $\rho=+0.57$ for both mean reward-model score and agreement rate. Self-BLEU, first-token entropy, and mean log-probability fall between these values.}

\revision{\textbf{Matched comparison.} On the same 50 configurations and with identical held-out folds, the compact predictor reaches $\rho=0.890$, and agreement rate alone reaches $0.57$ (\cref{tab:matched-screen}). The paired gain in correlation is $+0.31_{\pm .29}$, with a 95\% cluster-bootstrap interval above zero (\appref{app:bootstrap}). Mean answer entropy alone reaches $\rho=0.889$.}

\begin{table}[t]
  \centering
  \caption{\revision{Matched LOSO comparison on 50 configurations under identical outer folds.}}
  \label{tab:matched-screen}
  \small
  \begin{revised}
  \begin{tabular}{lc}
    \toprule
    Predictor & Spearman $\rho$ \\
    \midrule
    Compact predictor & $0.890$ \\
    Mean answer entropy & $0.889$ \\
    Single agreement rate & $0.57$ \\
    \bottomrule
  \end{tabular}
  \end{revised}
\end{table}

\revision{\textbf{Permutation test.} For the 18-feature ridge on the earlier 32-configuration set, the observed correlation exceeds the 97.5th percentile of the permutation distribution at each temperature ($p=0.002$, the smallest value resolved by the test). See \appref{app:permnull} for the full results.}

\subsection{\revision{Screening in practice}}
\label{sec:experiments:utility}

\revision{We use the earlier 18-feature ridge on 32 configurations. We average predicted and measured gains across temperatures for each configuration before ranking. At $K=5$, mean precision is $0.90$, and the selected set exactly matches the true top five in $64\%$ of bootstrap resamples (\cref{tab:precisionk}).}

\begin{figure}[t]
  \centering
  \includegraphics[width=\columnwidth]{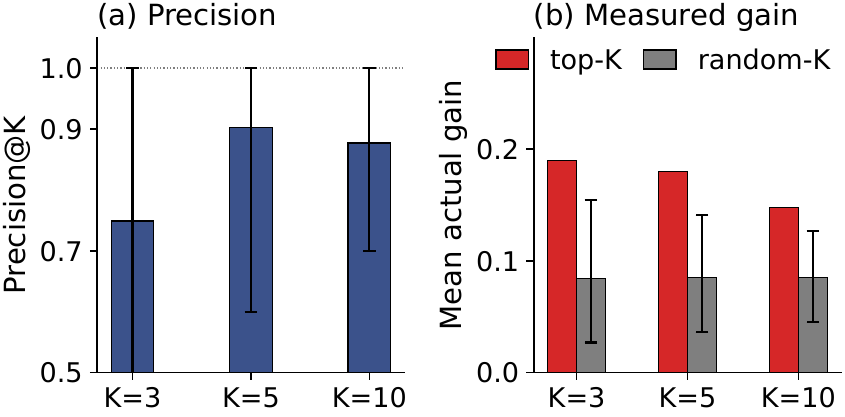}
  \caption{\revision{Precision and measured gain for configurations selected by predicted gain. We estimate precision by comparing the selected set with the measured top $K$ over 2,000 bootstrap resamples of checkpoint and domain groups. The right panel compares the gain of the selected configurations with the gain of a random set of size $K$.}}
  \label{fig:precisionk}
\end{figure}

\revision{\textbf{Gain over random selection.} The five highest-ranked configurations have a mean measured gain of $+0.18$, compared with $+0.09$ for five randomly selected configurations. The paired difference is $+0.10_{\pm .05}$, with a cluster-bootstrap confidence interval above zero. \Cref{fig:precisionk} shows the distributions of precision and gain across resamples.}

\begin{table}[h]
  \centering
  \caption{\revision{Mean measured gain for the top $K$ configurations by predicted gain and for $K$ randomly selected configurations. Gains are averaged over temperatures before ranking. We resample checkpoint and domain groups 2,000 times and draw the random set within each resample. The paired difference is $\Delta$. Precision and exact match compare the predicted and measured top $K$ within each resample.}}
  \label{tab:precisionk}
  \small
  \setlength{\tabcolsep}{3pt}
  \begin{tabular}{cccccc}
    \toprule
    $K$ & top-$K$ & random-$K$ & $\Delta$ & \revision{Precision} & \shortstack{\revision{Exact}\\\revision{match}} \\
    \midrule
    3  & $+0.19$ & $+0.09_{\pm .10}$ & $+0.11_{\pm .07}$ & $0.75$ & $43\%$ \\
    5  & $+0.18$ & $+0.09_{\pm .07}$ & $+0.10_{\pm .05}$ & $\mathbf{0.90}$ & $\mathbf{64\%}$ \\
    10 & $+0.15$ & $+0.09_{\pm .04}$ & $+0.06_{\pm .03}$ & $0.88$ & $27\%$ \\
    \bottomrule
  \end{tabular}
  
\end{table}

\revision{\textbf{How many scored configurations are needed?} We repeat the evaluation 200 times, fitting the compact predictor on selected checkpoint and task-domain groups and testing on the remaining groups. Each fit uses at least 16 scored configurations, with a median of 19 across repetitions. The median test correlation is $\rho=0.897$ (interquartile range $[0.868,0.914]$). The original LOSO evaluation on the 56-configuration grid gives $\rho=0.895$.}

\revision{\textbf{When does screening pay off?} At about 30 L40S GPU-minutes per configuration (\cref{tab:compute-breakdown}), scoring 19 configurations for training costs about 570 GPU-minutes. That cost is recovered once screening avoids full scoring on about 19 later configurations. Both workflows still generate the same answers, and configurations selected for full evaluation still need reward-model scores.}

\begin{table}[!htbp]
  \centering
  \caption{\revision{Measured time per configuration for screening and full evaluation. GPU times are for one L40S.}}
  \label{tab:compute-breakdown}
  \small
  \setlength{\tabcolsep}{4pt}
  \begin{revised}
  \begin{tabular}{@{}lll@{}}
    \toprule
    Stage & Time & Used by \\
    \midrule
    Candidate generation & 6--12 GPU-min & Both \\
    Skywork scoring & $\sim$30 GPU-min & Full evaluation \\
    Feature extraction & $\sim$3 CPU-min & Screen \\
    Ridge inference & CPU seconds & Screen \\
    \bottomrule
  \end{tabular}
  \end{revised}
\end{table}

\begin{samepage}
\subsection{\revision{Feature recurrence and entropy ablation}}
\label{sec:experiments:stability}

\revision{To see which features a sparse model keeps, we draw 500 bootstrap resamples of the matched 50 configurations and fit a Lasso to the gains in each \citep{Meinshausen2010Stability}, keeping all copies and temperatures of a configuration in the same fold when choosing the penalty. Five candidate features are selected in at least 80\% of fits. The feature that records when the first correct sample appears is selected in 78.6\% of fits. \appref{app:exp-stability} gives all selection frequencies. This analysis describes which features the Lasso selects. The ridge results use the same four features in every fold.}

\revision{\textbf{Removing the median entropy feature.} We keep the same 56 configurations and the same grouped inner and outer folds. Removing median per-prompt answer entropy lowers the correlation from $0.898$ to $0.827$. The baseline in \cref{tab:matched-screen} averages per-prompt entropy, whereas the feature removed here takes the median.}
\end{samepage}

\subsection{Robustness}
\label{sec:experiments:robustness}

\revision{We test five changes to the evaluation: the number of candidates, the gain function, the held-out post-training method, the prompts, and the reward model.}

\revision{\textbf{Across candidate counts.} We fit the 18-feature ridge to $\mathrm{BoN}@k' - \mathrm{pass}@1$ on 32 configurations for $k'\in\{2,4,8,16,32,64\}$, keeping the predictor inputs and $\mathrm{pass}@1$ fixed. Correlations are positive at every tested count and temperature (\cref{tab:k-sweep-app}).}

\revision{\textbf{Across gain functions.} We fit the 18-feature ridge to each of the four gains based on verifier selection on the same 32 configurations. Correlations range from $0.83$ to $0.88$. With majority voting, the correlation is $0.00_{\pm .46}$ (\cref{tab:gain-robustness}).}

\revision{\textbf{Across post-training methods.} We train the compact predictor on five post-training methods and evaluate it on the held-out sixth, repeating this for each method. Spearman correlations range from $0.78$ to $0.94$, and every cluster-bootstrap confidence interval excludes zero (\cref{tab:per-method}).}

\revision{\textbf{Across prompt sets.} We extract features from 64 samples per prompt on MATH500 \citep{Hendrycks2021Math, Lightman2023Lets} and use Skywork to measure gains. The four-feature ridge fitted on the original math pool reaches $\rho=0.79$ on MATH500 without refitting ($p<10^{-4}$).}

\revision{\textbf{Across reward models.} Both reward models see the same candidates and share the same $\mathrm{pass}@1$ baseline. We select the best-of-$N$ answer with ArmoRM-Llama3-8B \citep{Wang2024ArmoRM} and test whether the Skywork-trained predictor can rank the resulting gains. On the same 56 configurations, it reaches $\rho=0.85_{\pm .15}$ without refitting or tuning on ArmoRM gains. When we fit and evaluate within each reward model, the correlations are $0.81_{\pm .19}$ for ArmoRM and $0.90_{\pm .13}$ for Skywork (\cref{tab:cross-rm-target}).}

\begin{table}[!htbp]
  \centering
  \caption{\revision{Reward-model transfer on the same 56 configurations. Subscripts give 95\% cluster-bootstrap half-widths (\appref{app:bootstrap}).}}
  \label{tab:cross-rm-target}
  \small
  \setlength{\tabcolsep}{4pt}
  \begin{revised}
  \begin{tabular*}{\columnwidth}{@{\extracolsep{\fill}}llc@{}}
    \toprule
    Fit using gains from & Evaluated against & LOSO $\rho$ \\
    \midrule
    Skywork & Skywork & $0.90_{\pm .13}$ \\
    ArmoRM & ArmoRM & $0.81_{\pm .19}$ \\
    Skywork & ArmoRM & $0.85_{\pm .15}$ \\
    \bottomrule
  \end{tabular*}
  \end{revised}
\end{table}

\begin{samepage}
\subsection{Operating conditions}
\label{sec:experiments:failures}

\revision{We examine prediction errors at high sampling temperatures and when holding out programming tasks. We also compare estimates from individual temperatures with those from the joint ridge regression.}
\par
\end{samepage}

\revision{\textbf{Errors at high temperatures.} In the earlier 18-feature analysis, the largest LOSO errors occur for SFT and GRPO configurations at $T=0.7$ or $1.0$ (\cref{tab:adversarial-list}). Few samples agree, yet the reward model can still select a rare correct answer. Features such as majority-fraction spread and the position of the first correct sample describe the candidate answers. The reward model's scores determine which candidate is selected.}

\revision{\textbf{Holding out the programming domain.} In the earlier 18-feature analysis, holding out all programming configurations gives $\rho=-0.56$ (\appref{app:exp-failures}). For programming tasks, answer agreement compares function names and argument lists. Programs with the same signature can behave differently. A separate test on HumanEval uses fresh samples from prompts also in the original programming pool and gives $\rho=0.35$ (\appref{app:crossdata-bon}).}

\begingroup
\widowpenalty=10000
\revision{\textbf{Pooling temperatures.} We compare fits using one, two, or three temperatures on the same 32 configurations (\cref{tab:temperature-pooling}). The joint fit uses the same feature coefficients at every temperature and includes temperature as a covariate. It reaches $\rho=0.91$. The fit at $T=0.7$ alone reaches $0.72$, while the fits at $T=0.3$ and $1.0$ each reach at least $0.90$. The strongest pair, $T=0.3$ and $1.0$, gives $\rho=0.92$. For the main evaluation, we use the joint fit to predict gains at all three temperatures.}
\par
\endgroup

\begin{table}[!htbp]
  \centering
  \caption{\revision{Rank correlation when the compact predictor is fit and evaluated using different sampling temperatures on the same 32 configurations.}}
  \label{tab:temperature-pooling}
  \small
  \begin{revised}
  \begin{tabular}{lc}
    \toprule
    Temperatures used & LOSO $\rho$ \\
    \midrule
    $0.3$ & $0.90$ \\
    $0.7$ & $0.72$ \\
    $1.0$ & $0.91$ \\
    $0.3,\ 0.7$ & $0.89$ \\
    $0.3,\ 1.0$ & $0.92$ \\
    $0.7,\ 1.0$ & $0.88$ \\
    $0.3,\ 0.7,\ 1.0$ & $0.91$ \\
    \bottomrule
  \end{tabular}
  \end{revised}
\end{table}

\begin{samepage}
\section{Conclusion}
\label{sec:conclusion}

\revision{We developed a predictor of verifier-selected best-of-$N$ gains using statistics from a labeled validation set. The predictor recovered the gain ranking well on the combined math and programming evaluation set. We can use this ranking to screen configurations after generating candidates and reserve full reward-model scoring for those selected for evaluation.}
\par
\end{samepage}

\section*{Limitations}
\label{sec:limitations}

\revision{The present study focuses on verifier-selected best-of-$N$ for math and programming benchmarks.} Natural extensions include open-ended generation, tool-augmented tasks, and future model families whose output distributions may differ from those studied here. The predictor is intended as a pre-deployment screening tool for \revision{comparing configurations on held-out prompts. Final deployment decisions require task-specific evaluation.} \revision{The number of scored configurations needed for training may vary with the reward model and task domain. Future work could hold the base model and training objective fixed and test whether the mix of fine-tuning data \citep{zhang2026pickmixturesamplescombinatorial} changes best-of-$N$ gains.}

\bibliography{references}

@article{Austin2021Program,
  title={Program Synthesis with Large Language Models},
  author={Austin, Jacob and Odena, Augustus and Nye, Maxwell and Bosma, Maarten and Michalewski, Henryk and Dohan, David and Jiang, Ellen and Cai, Carrie and Terry, Michael and Le, Quoc and Sutton, Charles},
  journal={arXiv preprint arXiv:2108.07732},
  year={2021}
}

@article{Chen2021Evaluating,
  title={Evaluating Large Language Models Trained on Code},
  author={Chen, Mark and Tworek, Jerry and Jun, Heewoo and Yuan, Qiming and Pinto, Henrique Ponde de Oliveira and Kaplan, Jared and Edwards, Harri and Burda, Yuri and Joseph, Nicholas and Brockman, Greg and Ray, Alex and Puri, Raul and Krueger, Gretchen and Petrov, Michael and Khlaaf, Heidy and Sastry, Girish and Mishkin, Pamela and Chan, Brooke and Gray, Scott and Ryder, Nick and Pavlov, Mikhail and Power, Alethea and Kaiser, Lukasz and Bavarian, Mohammad and Winter, Clemens and Tillet, Philippe and Such, Felipe Petroski and Cummings, Dave and Plappert, Matthias and Chantzis, Fotios and Barnes, Elizabeth and Herbert-Voss, Ariel and Guss, William Hebgen and Nichol, Alex and Paino, Alex and Tezak, Nikolas and Tang, Jie and Babuschkin, Igor and Balaji, Suchir and Jain, Shantanu and Saunders, William and Hesse, Christopher and Carr, Andrew N. and Leike, Jan and Achiam, Josh and Misra, Vedant and Morikawa, Evan and Radford, Alec and Knight, Matthew and Brundage, Miles and Murati, Mira and Mayer, Katie and Welinder, Peter and McGrew, Bob and Amodei, Dario and McCandlish, Sam and Sutskever, Ilya and Zaremba, Wojciech},
  journal={arXiv preprint arXiv:2107.03374},
  year={2021}
}

@inproceedings{Azaria2023Internal,
  title={The Internal State of an {LLM} Knows When It's Lying},
  author={Azaria, Amos and Mitchell, Tom},
  booktitle={Findings of the Association for Computational Linguistics: EMNLP 2023},
  year={2023},
  pages={967--976},
  url={https://aclanthology.org/2023.findings-emnlp.68/}
}

@article{Brown2024Large,
  title={Large Language Monkeys: Scaling Inference Compute with Repeated Sampling},
  author={Brown, Bradley and Juravsky, Jordan and Ehrlich, Ryan and Clark, Ronald and Le, Quoc V and R\'e, Christopher and Mirhoseini, Azalia},
  journal={arXiv preprint arXiv:2407.21787},
  year={2024}
}

@article{Burnell2023Revealing,
  title={Revealing the Structure of Language Model Capabilities},
  author={Burnell, Ryan and Hao, Han and Conway, Andrew R. A. and Hern{\'a}ndez-Orallo, Jos{\'e}},
  journal={arXiv preprint arXiv:2306.10062},
  year={2023}
}

@inproceedings{Burns2023Discovering,
  title={Discovering Latent Knowledge in Language Models Without Supervision},
  author={Burns, Collin and Ye, Haotian and Klein, Dan and Steinhardt, Jacob},
  booktitle={International Conference on Learning Representations (ICLR)},
  year={2023}
}

@article{Cobbe2021Training,
  title={Training Verifiers to Solve Math Word Problems},
  author={Cobbe, Karl and Kosaraju, Vineet and Bavarian, Mohammad and Chen, Mark and Jun, Heewoo and Kaiser, Lukasz and Plappert, Matthias and Tworek, Jerry and Hilton, Jacob and Nakano, Reiichiro and Hesse, Christopher and Schulman, John},
  journal={arXiv preprint arXiv:2110.14168},
  year={2021}
}

@inproceedings{Ethayarajh2024KTO,
  title={Model Alignment as Prospect Theoretic Optimization},
  author={Ethayarajh, Kawin and Xu, Winnie and Muennighoff, Niklas and Jurafsky, Dan and Kiela, Douwe},
  year={2024},
  booktitle={Proceedings of the 41st International Conference on Machine Learning},
  series={Proceedings of Machine Learning Research},
  volume={235},
  pages={12634--12651},
  publisher={PMLR},
  url={https://proceedings.mlr.press/v235/ethayarajh24a.html}
}

@inproceedings{Hendrycks2021Math,
  title={Measuring Mathematical Problem Solving With the {MATH} Dataset},
  author={Hendrycks, Dan and Burns, Collin and Kadavath, Saurav and Arora, Akul and Basart, Steven and Tang, Eric and Song, Dawn and Steinhardt, Jacob},
  booktitle={Neural Information Processing Systems (NeurIPS) Datasets and Benchmarks Track},
  year={2021}
}

@inproceedings{Hoffmann2022Training,
  title={Training Compute-Optimal Large Language Models},
  author={Hoffmann, Jordan and Borgeaud, Sebastian and Mensch, Arthur and Buchatskaya, Elena and Cai, Trevor and Rutherford, Eliza and de Las Casas, Diego and Hendricks, Lisa Anne and Welbl, Johannes and Clark, Aidan and Hennigan, Thomas and Noland, Eric and Millican, Katherine and van den Driessche, George and Damoc, Bogdan and Guy, Aurelia and Osindero, Simon and Simonyan, Kar{\'e}n and Elsen, Erich and Vinyals, Oriol and Rae, Jack and Sifre, Laurent},
  year={2022},
  booktitle={Advances in Neural Information Processing Systems},
  volume={35},
  pages={30016--30030},
  doi={10.52202/068431-2176}
}

@inproceedings{Holtzman2020Nucleus,
  title={The Curious Case of Neural Text Degeneration},
  author={Holtzman, Ari and Buys, Jan and Du, Li and Forbes, Maxwell and Choi, Yejin},
  booktitle={International Conference on Learning Representations (ICLR)},
  year={2020}
}

@inproceedings{Hong2024ORPO,
  title={{ORPO}: Monolithic Preference Optimization without Reference Model},
  author={Hong, Jiwoo and Lee, Noah and Thorne, James},
  booktitle={Proceedings of the 2024 Conference on Empirical Methods in Natural Language Processing},
  pages={11170--11189},
  year={2024}
}

@article{Jiang2021How,
  title={How Can We Know When Language Models Know? On the Calibration of Language Models for Question Answering},
  author={Jiang, Zhengbao and Araki, Jun and Ding, Haibo and Neubig, Graham},
  journal={Transactions of the Association for Computational Linguistics},
  volume={9},
  pages={962--977},
  year={2021}
}

@article{Kadavath2022Language,
  title={Language Models (Mostly) Know What They Know},
  author={Kadavath, Saurav and Conerly, Tom and Askell, Amanda and Henighan, Tom and Drain, Dawn and Perez, Ethan and Schiefer, Nicholas and Hatfield-Dodds, Zac and DasSarma, Nova and Tran-Johnson, Eli and Johnston, Scott and El-Showk, Sheer and Jones, Andy and Elhage, Nelson and Hume, Tristan and Chen, Anna and Bai, Yuntao and Bowman, Sam and Fort, Stanislav and Ganguli, Deep and Hernandez, Danny and Jacobson, Josh and Kernion, Jackson and Kravec, Shauna and Lovitt, Liane and Ndousse, Kamal and Olsson, Catherine and Ringer, Sam and Amodei, Dario and Brown, Tom and Clark, Jack and Joseph, Nicholas and Mann, Ben and McCandlish, Sam and Olah, Chris and Kaplan, Jared},
  journal={arXiv preprint arXiv:2207.05221},
  year={2022}
}

@article{Kaplan2020Scaling,
  title={Scaling Laws for Neural Language Models},
  author={Kaplan, Jared and McCandlish, Sam and Henighan, Tom and Brown, Tom B. and Chess, Benjamin and Child, Rewon and Gray, Scott and Radford, Alec and Wu, Jeffrey and Amodei, Dario},
  journal={arXiv preprint arXiv:2001.08361},
  year={2020}
}

@inproceedings{Kirk2024Understanding,
  title={Understanding the Effects of {RLHF} on {LLM} Generalisation and Diversity},
  author={Kirk, Robert and Mediratta, Ishita and Nalmpantis, Christoforos and Luketina, Jelena and Hambro, Eric and Grefenstette, Edward and Raileanu, Roberta},
  booktitle={International Conference on Learning Representations (ICLR)},
  year={2024}
}

@inproceedings{Kwon2023Efficient,
  title={Efficient Memory Management for Large Language Model Serving with {PagedAttention}},
  author={Kwon, Woosuk and Li, Zhuohan and Zhuang, Siyuan and Sheng, Ying and Zheng, Lianmin and Yu, Cody Hao and Gonzalez, Joseph E. and Zhang, Hao and Stoica, Ion},
  booktitle={Proceedings of the 29th Symposium on Operating Systems Principles},
  year={2023},
  pages={611--626},
  doi={10.1145/3600006.3613165}
}

@inproceedings{Lightman2023Lets,
  title={Let's Verify Step by Step},
  author={Lightman, Hunter and Kosaraju, Vineet and Burda, Yuri and Edwards, Harrison and Baker, Bowen and Lee, Teddy and Leike, Jan and Schulman, John and Sutskever, Ilya and Cobbe, Karl},
  booktitle={International Conference on Learning Representations (ICLR)},
  year={2024}
}

@article{Liu2024Skywork,
  title={Skywork-Reward: Bag of Tricks for Reward Modeling in {LLMs}},
  author={Liu, Chris Yuhao and Zeng, Liang and Liu, Jiacai and Yan, Rui and He, Jujie and Wang, Chaojie and Yan, Shuicheng and Liu, Yang and Zhou, Yahui},
  journal={arXiv preprint arXiv:2410.18451},
  year={2024}
}

@inproceedings{Wang2024ArmoRM,
  title={Interpretable Preferences via Multi-Objective Reward Modeling and Mixture-of-Experts},
  author={Wang, Haoxiang and Xiong, Wei and Xie, Tengyang and Zhao, Han and Zhang, Tong},
  booktitle={Findings of the Association for Computational Linguistics: EMNLP 2024},
  year={2024},
  pages={10582--10592},
  url={https://aclanthology.org/2024.findings-emnlp.620/}
}

@article{McKenzie2023Inverse,
  title={Inverse Scaling: When Bigger Isn't Better},
  author={Ian R. McKenzie and Alexander Lyzhov and Michael Martin Pieler and Alicia Parrish and Aaron Mueller and Ameya Prabhu and Euan McLean and Xudong Shen and Joe Cavanagh and Andrew George Gritsevskiy and Derik Kauffman and Aaron T. Kirtland and Zhengping Zhou and Yuhui Zhang and Sicong Huang and Daniel Wurgaft and Max Weiss and Alexis Ross and Gabriel Recchia and Alisa Liu and Jiacheng Liu and Tom Tseng and Tomasz Korbak and Najoung Kim and Samuel R. Bowman and Ethan Perez},
  journal={Transactions on Machine Learning Research},
  year={2023},
  url={https://openreview.net/forum?id=DwgRm72GQF}
}

@article{Meinshausen2010Stability,
  title={Stability Selection},
  author={Meinshausen, Nicolai and B\"uhlmann, Peter},
  journal={Journal of the Royal Statistical Society: Series B (Statistical Methodology)},
  volume={72},
  number={4},
  pages={417--473},
  year={2010}
}

@inproceedings{Meng2024SimPO,
  title={{SimPO}: Simple Preference Optimization with a Reference-Free Reward},
  author={Meng, Yu and Xia, Mengzhou and Chen, Danqi},
  booktitle={Advances in Neural Information Processing Systems},
  year={2024},
  volume={37},
  pages={124198--124235},
  doi={10.52202/079017-3946}
}

@inproceedings{Ouyang2022Training,
  title={Training Language Models to Follow Instructions with Human Feedback},
  author={Ouyang, Long and Wu, Jeff and Jiang, Xu and Almeida, Diogo and Wainwright, Carroll L. and Mishkin, Pamela and Zhang, Chong and Agarwal, Sandhini and Slama, Katarina and Ray, Alex and Schulman, John and Hilton, Jacob and Kelton, Fraser and Miller, Luke and Simens, Maddie and Askell, Amanda and Welinder, Peter and Christiano, Paul and Leike, Jan and Lowe, Ryan},
  volume={35},
  year={2022},
  booktitle={Advances in Neural Information Processing Systems},
  pages={27730--27744},
  doi={10.52202/068431-2011}
}

@inproceedings{Rafailov2023DPO,
  title={Direct Preference Optimization: Your Language Model is Secretly a Reward Model},
  author={Rafailov, Rafael and Sharma, Archit and Mitchell, Eric and Ermon, Stefano and Manning, Christopher D. and Finn, Chelsea},
  booktitle={Advances in Neural Information Processing Systems},
  year={2023},
  volume={36},
  pages={53728--53741},
  doi={10.52202/075280-2338}
}

@inproceedings{Ruan2024Observational,
  title={Observational Scaling Laws and the Predictability of Language Model Performance},
  author={Ruan, Yangjun and Maddison, Chris J. and Hashimoto, Tatsunori},
  year={2024},
  booktitle={Advances in Neural Information Processing Systems},
  volume={37},
  pages={15841--15892},
  doi={10.52202/079017-0506}
}

@inproceedings{Schaeffer2023Are,
  title={Are Emergent Abilities of Large Language Models a Mirage?},
  author={Schaeffer, Rylan and Miranda, Brando and Koyejo, Sanmi},
  booktitle={Advances in Neural Information Processing Systems},
  year={2023},
  volume={36},
  pages={55565--55581},
  doi={10.52202/075280-2425}
}

@article{Shao2024DeepSeekMath,
  title={{DeepSeekMath}: Pushing the Limits of Mathematical Reasoning in Open Language Models},
  author={Shao, Zhihong and Wang, Peiyi and Zhu, Qihao and Xu, Runxin and Song, Junxiao and Bi, Xiao and Zhang, Haowei and Zhang, Mingchuan and Li, Y. K. and Wu, Y. and Guo, Daya},
  journal={arXiv preprint arXiv:2402.03300},
  year={2024}
}

@inproceedings{Snell2024Scaling,
  title={Scaling {LLM} Test-Time Compute Optimally Can be More Effective than Scaling Parameters for Reasoning},
  author={Snell, Charlie and Lee, Jaehoon and Xu, Kelvin and Kumar, Aviral},
  year={2025},
  booktitle={International Conference on Learning Representations},
  pages={10131--10165},
  url={https://proceedings.iclr.cc/paper_files/paper/2025/hash/1b623663fd9b874366f3ce019fdfdd44-Abstract-Conference.html}
}

@article{Uesato2022Solving,
  title={Solving Math Word Problems with Process- and Outcome-Based Feedback},
  author={Uesato, Jonathan and Kushman, Nate and Kumar, Ramana and Song, Francis and Siegel, Noah and Wang, Lisa and Creswell, Antonia and Irving, Geoffrey and Higgins, Irina},
  journal={arXiv preprint arXiv:2211.14275},
  year={2022}
}

@inproceedings{Wang2022Self,
  title={Self-Consistency Improves Chain of Thought Reasoning in Language Models},
  author={Wang, Xuezhi and Wei, Jason and Schuurmans, Dale and Le, Quoc V. and Chi, Ed H. and Narang, Sharan and Chowdhery, Aakanksha and Zhou, Denny},
  booktitle={International Conference on Learning Representations (ICLR)},
  year={2023}
}

@inproceedings{Wu2024Inference,
  title={Inference Scaling Laws: An Empirical Analysis of Compute-Optimal Inference for {LLM} Problem-Solving},
  author={Wu, Yangzhen and Sun, Zhiqing and Li, Shanda and Welleck, Sean and Yang, Yiming},
  year={2025},
  booktitle={International Conference on Learning Representations},
  pages={55824--55845},
  url={https://proceedings.iclr.cc/paper_files/paper/2025/hash/8c3caae2f725c8e2a55ecd600563d172-Abstract-Conference.html}
}

@inproceedings{Zhu2018Texygen,
  title={{Texygen}: A Benchmarking Platform for Text Generation Models},
  author={Zhu, Yaoming and Lu, Sidi and Zheng, Lei and Guo, Jiaxian and Zhang, Weinan and Wang, Jun and Yu, Yong},
  booktitle={Proceedings of the 41st International ACM SIGIR Conference on Research and Development in Information Retrieval},
  pages={1097--1100},
  year={2018},
  doi={10.1145/3209978.3210080}
}

@misc{zhang2026posttrainingrecipemodelfamily,
      title={Post-Training Recipe, More Than Model Family, Shapes Multi-Agent {LLM} Conversational Behavior},
      author={Luyang Zhang and Jialu Wang and Fei Xue and Yi-Yun Chu},
      year={2026},
      eprint={2606.20632},
      archivePrefix={arXiv},
      primaryClass={cs.CL},
      url={https://arxiv.org/abs/2606.20632},
}

@misc{zhang2026spectrarevealingspectrumuser,
      title={{SPECTRA}: Revealing the Full Spectrum of User Preferences via Distributional {LLM} Inference},
      author={Luyang Zhang and Jialu Wang and Shichao Zhu and Beibei Li and Zhongcun Wang and Guangmou Pan and Yang Song},
      year={2026},
      eprint={2509.24189},
      archivePrefix={arXiv},
      primaryClass={cs.CL},
      url={https://arxiv.org/abs/2509.24189},
}

@misc{zhang2026attributionpatchingliesdiagnosis,
      title={When Attribution Patching Lies: Diagnosis and a Second-Order Correction},
      author={Luyang Zhang and Jialu Wang},
      year={2026},
      eprint={2606.09899},
      archivePrefix={arXiv},
      primaryClass={cs.LG},
      url={https://arxiv.org/abs/2606.09899},
}

@misc{zhang2026correctingattributionpatchingexact,
      title={Correcting Attribution Patching with a Few Exact Measurements},
      author={Luyang Zhang and Minxing Zheng and Jialu Wang},
      year={2026},
      eprint={2610.27393},
      archivePrefix={arXiv},
      primaryClass={cs.AI},
      url={https://arxiv.org/abs/2610.27393},
}

@misc{zhang2026pickmixturesamplescombinatorial,
      title={Pick the Mixture, Not the Samples: Combinatorial Data Selection for {LLMs}},
      author={Luyang Zhang and Jialu Wang and Beibei Li},
      year={2026},
      eprint={2610.14028},
      archivePrefix={arXiv},
      primaryClass={cs.LG},
      url={https://arxiv.org/abs/2610.14028},
}

@article{Hoeffding1963Probability,
  author    = {Hoeffding, Wassily},
  title     = {Probability Inequalities for Sums of Bounded Random Variables},
  journal   = {Journal of the American Statistical Association},
  year      = {1963},
  volume    = {58},
  number    = {301},
  pages     = {13--30},
  doi       = {10.1080/01621459.1963.10500830},
  url       = {https://doi.org/10.1080/01621459.1963.10500830}
}

@inproceedings{Hsu2012RandomDesign,
  author    = {Hsu, Daniel and Kakade, Sham M. and Zhang, Tong},
  title     = {Random Design Analysis of Ridge Regression},
  booktitle = {Proceedings of the 25th Annual Conference on Learning Theory},
  series    = {Proceedings of Machine Learning Research},
  year      = {2012},
  volume    = {23},
  pages     = {9.1--9.24},
  publisher = {PMLR},
  url       = {https://proceedings.mlr.press/v23/hsu12.html}
}

\appendix
\crefname{section}{Appendix}{Appendices}
\Crefname{section}{Appendix}{Appendices}
\sloppy
\emergencystretch=3em

\section{Detailed setups}
\label{app:setups}
\label{app:reproducibility}

\subsection{Sampling and inference}
\revision{We use vLLM \citep{Kwon2023Efficient} to generate 64 samples per prompt at each of three temperatures ($T\in\{0.3,0.7,1.0\}$). For the main evaluation, we use top-$p=0.95$ and a 512-token limit. We take the first 200 problems in the GSM8K test split \citep{Cobbe2021Training}. The programming pool contains all 164 HumanEval problems \citep{Chen2021Evaluating} and the first 36 problems in the MBPP test split \citep{Austin2021Program}. Configurations in the same domain share the same prompts and ordering.}

\subsection{Evaluation-set sizes}
\revision{The frozen evaluation for the compact predictor in \cref{tab:headline} contains 56 configurations (40 math and 16 code), each evaluated at three temperatures, for 168 rows. LOSO holds out one checkpoint and task-domain group at a time. The two multi-feature baselines in \cref{tab:headline} and the comparison in \cref{tab:matched-screen} use 50 configurations with complete features. The single-feature baselines use available values from that set. First-token entropy is available on 32 configurations.}
\revision{The evaluated checkpoints include Qwen2.5-3B/7B-Instruct, Llama-3.2-3B-Instruct, Llama-3.1-8B-Instruct, and gemma-2-2B/9B-it. \Cref{fig:calksweep}a plots the 168 held-out predictions for the compact predictor. The candidate-count sweep in \cref{fig:calksweep}b uses the earlier 32-configuration set (23 math and 9 code).}

\subsection{Artifact use and data handling}
\revision{We use public benchmark artifacts, including GSM8K, HumanEval, MBPP, and MATH500,} public model and reward-model artifacts (the base-model families in \cref{sec:experiments:setup}, Skywork-Reward-Llama-3.1-8B, and ArmoRM-Llama3-8B), and the vLLM inference library. We cite the creators of these artifacts in the main text and this appendix. Our use is limited to research evaluation of model configurations; we do not redistribute benchmark datasets, model weights, reward-model weights, or raw generated completions, and any released code or derived tables are intended to require users to obtain the original artifacts under their own licenses or terms. \revision{This study involved no human subjects or annotators.} We use generated samples only for aggregate feature extraction and evaluation statistics.

\subsection{Reward models}
The primary reward model is Skywork-Reward-Llama-3.1-8B \citep{Liu2024Skywork} (8B parameters, Llama-3.1 architecture). The target-level cross-reward-model robustness check (\cref{tab:cross-rm-target}) uses ArmoRM-Llama3-8B \citep{Wang2024ArmoRM} (8B parameters, Llama-3 architecture with a multi-objective mixture-of-experts head), independently trained by a different group on different preference data.

\subsection{Predictor configuration}
\label{app:predictor-configuration}
\revision{For \cref{tab:headline}, we standardize features on the outer training fold and use \texttt{RidgeCV} with five inner folds over temperature rows. In the grouped rerun, we use the same penalty grid, hold out whole checkpoint and task-domain groups in the inner folds, and fit the scaler separately on the training portion of each inner fold. The grid is $\alpha\in\{10^{-3},10^{-2},10^{-1},1,10,10^2\}$. We center temperature at $T_0=0.7$.}

\subsection{Cluster bootstrap}
\label{app:bootstrap}
\revision{For the LOSO intervals in \cref{tab:headline}, we resample whole checkpoint and domain groups. For LOO, we resample individual configurations. Both procedures keep all temperatures of each sampled unit together. We recompute Spearman correlation over 500 resamples and use the 2.5th and 97.5th percentiles as the interval bounds. A paired bootstrap uses the same draws for the two predictions being compared.}

\revision{We write $x_{\pm h}$ for the observed value $x$ and half the distance $h$ between the percentile bounds. The exact 95\% intervals are $[0.03,0.61]$ for the paired difference between the compact predictor and agreement rate alone, $[-0.03,+0.16]$ for the paired difference against the output-distribution ridge, and $[0.62,0.92]$ for Skywork-to-ArmoRM transfer.}

\subsection{\revision{Feature recurrence bootstrap}}
\revision{We draw 500 bootstrap resamples from the same set of 50 configurations. Each Lasso fit standardizes features using only its training fold and uses grouped inner cross-validation. If a configuration is drawn more than once, all copies stay in the same fold.}

\subsection{Permutation null}
\revision{For the permutation test, we fit the 18-feature predictor on the earlier set of 32 configurations. We shuffle gains across configurations 500 times, keeping all three temperatures of a configuration together, and refit in every LOSO fold. The smallest one-sided $p$-value is $1/(500+1)$, reported as $0.002$.}

\subsection{Compute}
\revision{Inference and post-training ran on L40S 48GB GPUs. Feature extraction and ridge fitting use CPUs after candidate generation.}

\subsection{Cost-value comparison}
\revision{Both workflows generate $k=64$ candidates for each of $P=200$ prompts at three temperatures. For full best-of-$N$ evaluation, Skywork scores all $P\times k\times 3=38{,}400$ prompt--completion pairs. We timed this step on an L40S 48GB at batch size 32 in bf16 (\cref{tab:compute-breakdown}). The break-even calculation charges for the initial scored configurations and credits only later configurations whose full scoring is skipped. A 19-configuration training set costs about 570 L40S GPU-minutes at the measured scoring rate.}

\revision{In each of the 200 repetitions, we fit the compact predictor on whole checkpoint and task-domain groups and test on the remaining groups. Every fit uses at least 16 scored configurations, with a median of 19 across repetitions.}

\section{Framework details}
\label{app:framework}

\revision{This section gives the feature catalog, the proof of \cref{thm:joint-bound}, and the details of the held-out extension.}

\subsection{Feature catalog}
\label{app:features}

\revision{The catalog contains 18 features. We compute each one from 64 sampled completions per prompt across 200 prompts at a fixed temperature.}

\begin{table*}[h]
  \centering
  \caption{\revision{Feature catalog. We count reasoning steps using rules specific to each domain. The position of the first correct sample requires gold validation answers.}}
  \label{tab:feature-catalog}
  \vspace{-3pt}
  \small
  \setlength{\tabcolsep}{6pt}
  \begin{tabular}{lll}
    \toprule
    Feature (code name) & Family & Description \\
    \midrule
    \texttt{agreement\_rate} & \revision{output distribution} & \revision{share of pairs with matching answers} \\
    \texttt{majority\_fraction} & \revision{output distribution} & avg.\ per-prompt modal fraction \\
    \texttt{self\_bleu} & \revision{output distribution} & \revision{4-gram overlap over sampled pairs} \\
    \texttt{uniq\_2gram\_ratio} & \revision{output distribution} & \revision{distinct bigrams / total bigrams} \\
    \texttt{embed\_sim} & \revision{output distribution} & pairwise cosine sim.\ of sample embeds \\
    \texttt{answer\_entropy} & \revision{output distribution} & Shannon entropy of extracted answers \\
    \texttt{mean\_logprob} & \revision{output distribution} & mean per-token log-prob \\
    \texttt{std\_logprob} & \revision{output distribution} & \revision{mean within-completion log-prob SD} \\
    \texttt{mean\_topK\_mass} & \revision{output distribution} & avg.\ top-$K$ token mass per step \\
    \texttt{seq\_lp\_spread} & \revision{output distribution} & cross-sample sequence log-prob spread \\
    \midrule
    \texttt{majority\_size\_std} & \revision{additional statistics} & cross-prompt SD of modal fraction \\
    \texttt{majority\_size\_min} & \revision{additional statistics} & min cross-prompt modal fraction \\
    \texttt{completion\_length\_variance} & \revision{additional statistics} & \revision{mean within-prompt length SD} \\
    \texttt{first\_correct\_sample\_position\_median} & \revision{additional statistics} & median first-correct index \\
    \texttt{per\_prompt\_answer\_entropy\_median} & \revision{additional statistics} & median per-prompt answer entropy \\
    \texttt{completion\_repetition\_4gram} & \revision{additional statistics} & 4-gram repetition rate \\
    \texttt{reasoning\_step\_count\_mean} & \revision{additional statistics} & domain-adaptive step count \\
    \revision{\texttt{chosen\_logprob\_disagreement\_pairwise}} & \revision{additional statistics} & \revision{pairwise chosen-token disagreement} \\
    \bottomrule
  \end{tabular}

\end{table*}

\revision{\textbf{Compact predictor inputs.} Majority-fraction spread is the standard deviation across prompts of the fraction of samples giving each prompt's most frequent answer. We compute Shannon entropy over extracted answers for each prompt. The compact predictor uses the median of these entropies. The output-distribution ridge includes the mean as one of its inputs, and \cref{tab:matched-screen} reports the mean as a standalone baseline. The joint ridge fits all its inputs together, so its rank correlation can differ from that of an individual input.}

\revision{For completion-length variation, we measure completion lengths in characters, compute their standard deviation within each prompt, and average across prompts. The implementation stores this statistic under the name \texttt{completion\_length\_variance}. For first-correct-sample position, we record the index of the first correct sample and take the median across prompts with at least one correct sample. This calculation uses the gold validation answers.}

\subsection{\revision{Proof of the joint concentration bound}}
\label{app:proof}

\revision{We prove \cref{thm:joint-bound}.} Recall that we wish to bound
\[
\bigl|\,\hat{\boldsymbol{\beta}}^{\!\top}\hat{\mathbf{x}}(c, T) - G(c, T)\,\bigr|
\]
for any Lipschitz gain function $G \in \mathcal{G}_L$ uniformly over all $m$ training configurations $c$ and all three temperatures $T$, with probability at least $1 - \delta$.

\textit{Decomposition.} By the triangle inequality,
\begin{align*}
\bigl|\,\hat{\boldsymbol{\beta}}^{\!\top}\hat{\mathbf{x}} - G(c,T)\,\bigr|
&\leq \underbrace{\bigl|\hat{\boldsymbol{\beta}}^{\!\top}(\hat{\mathbf{x}} - \mathbf{x})\bigr|}_{(A)\ \text{\revision{features}}} \\
&\quad + \underbrace{\bigl|(\hat{\boldsymbol{\beta}} - \boldsymbol{\beta}^\star)^{\!\top}\mathbf{x}\bigr|}_{(B)\ \text{coef.}} \\
&\quad + \underbrace{\bigl|\boldsymbol{\beta}^{\star\,\top}\mathbf{x} - G(c,T)\bigr|}_{(C)\ \text{approx.}}.
\end{align*}

\textit{Term (A).} By H\"older's inequality with the dual $(\ell^1, \ell^\infty)$ pair,
\[
\bigl|\hat{\boldsymbol{\beta}}^{\!\top}(\hat{\mathbf{x}} - \mathbf{x})\bigr|
\leq \sum_{j=1}^{d} |\hat\beta_j|\,|\hat{x}_j - x_j|.
\]
\revision{Let $\varepsilon_j$ bound the estimation error of feature $j$ at failure probability at most $\delta_{\text{feat}}/(3md)$ for each training configuration and temperature. A union bound over features, configurations, and temperatures gives, with probability at least $1-\delta_{\text{feat}}$,}
\[
|\hat{x}_j - x_j| \leq \varepsilon_j \qquad \text{for all } j = 1, \dots, d,
\]
\revision{simultaneously over the training configurations and temperatures. Summing the errors with weights $|\hat\beta_j|$ yields $E_\beta(\hat{\boldsymbol{\beta}})=\sum_j|\hat\beta_j|\varepsilon_j$. The radii must be expressed in the feature units used to fit the coefficients. For features bounded in $[0,1]$, $\varepsilon_j=1$ always gives a bound. A smaller radius for a median or a standard deviation needs an argument for that statistic. \appref{app:predictor-configuration} describes standardization for each evaluation protocol.}

\textit{Term (B).} By Cauchy--Schwarz,
\[
\bigl|(\hat{\boldsymbol{\beta}} - \boldsymbol{\beta}^\star)^{\!\top}\mathbf{x}\bigr|
\leq \|\hat{\boldsymbol{\beta}} - \boldsymbol{\beta}^\star\|_2 \cdot \|\mathbf{x}(c,T)\|_2.
\]
\revision{The held-out extension uses the coefficient error bound assumed in \cref{cor:heldout-transfer}.}

\textit{Term (C).} Lipschitzness alone does not imply that the gain function lies in the linear span of the selected features. We therefore keep the approximation residual explicit:
\[
\bigl|\boldsymbol{\beta}^{\star\,\top}\mathbf{x} - G(c,T)\bigr|
\leq
A_G^\star,
\]
where $A_G^\star := \sup_{c,T}|\boldsymbol{\beta}^{\star\top}\mathbf{x}(c,T)-G(c,T)|$ over the training configurations and temperatures. This term is the modeling error of the linear feature representation. \revision{The held-out experiments assess prediction accuracy, which includes this error.}

\revision{\textit{Target-side concentration.} The two accuracies are means of bounded values over $P$ independent prompts. By two-sided Hoeffding bounds and a union bound over the two means, $m$ configurations, and three temperatures, both estimation errors are at most $r=\sqrt{\log(12m/\delta_{\text{tgt}})/(2P)}$ simultaneously with probability at least $1-\delta_{\text{tgt}}$. With $\sigma_{\text{tgt}}=\sqrt{2}\,r$,}
\begin{align*}
\bigl\|(\widehat{\mathrm{BoN}@k}, \widehat{\mathrm{pass}@1}) &- (\mathrm{BoN}@k, \mathrm{pass}@1)\bigr\|_2 \\
&\leq \sigma_{\text{tgt}}.
\end{align*}
For any $G \in \mathcal{G}_L$, this implies $|\hat G(c,T)-G(c,T)| \leq L_G\,\sigma_{\text{tgt}}$.

\textit{Symmetric split.} Setting $\delta_{\text{tgt}} = \delta_{\text{feat}} = \delta/2$ and combining the feature, coefficient, approximation, and target-concentration bounds yields the statement of \cref{thm:joint-bound}. \qed

\subsection{\revision{Held-out extension}}
\label{app:bound-eval}

\revision{\Cref{cor:heldout-transfer} applies the feature and gain error bounds to an independent held-out configuration. It uses the approximation error $A_{G,\mathrm{pop}}$ over the configuration population. On the event that the coefficient error is at most $\eta_m$, Cauchy--Schwarz bounds its contribution by $R_x\eta_m$. The union bound gives the stated probability for one held-out configuration and its measured gain. To apply \cref{lem:rank-perturb} to population gains, we need a bound that holds simultaneously across the configurations being ranked. For a finite set, one can allocate the feature and population-to-measured gain error probabilities across that set and combine the errors by the triangle inequality.}

\revision{\textbf{Proof of \cref{lem:rank-perturb}.} A pair with a population gain gap greater than $2\Delta$ keeps its order. Thus at most $D=q_{\leq 2\Delta}m'(m'-1)/2$ pair orders can differ. Each adjacent swap changes the sum of absolute rank displacements by at most two, so this sum is at most $2D$. If $d_i$ is the rank displacement of configuration $i$, then $\sum_i d_i^2\leq (m'-1)2D$. Substituting into the Spearman formula gives $1-\hat\rho\leq 6q_{\leq 2\Delta}$.}

\section{Experiment extensions}
\label{app:experiments}

\revision{This section reports additional analyses for the experiments in \cref{sec:experiments}. \appref{app:setups} gives the evaluation setup.}

\subsection{Rank prediction}
\label{app:exp-rank}

\revision{\Cref{tab:temperature-pooling} reports the matched comparison of fits using different temperatures on 32 configurations. The permutation test below reports separate correlations by temperature from a single fit across all three temperatures.}

\paragraph{Permutation null.}
\label{app:permnull}
\revision{\Cref{tab:permnull-app} reports the 18-feature fit on the earlier 32-configuration set described in \appref{app:setups}. The observed correlations are computed separately by temperature. They are the same $k'=64$ correlations reported in \cref{tab:k-sweep-app}. The null columns give percentiles of the permutation distribution.}

\begin{table}[h]
  \centering
  \caption{\revision{Permutation test for the 18-feature predictor at $k=64$ on 32 configurations. The null interval contains the central 95\% of the permutation correlations.}}
  \label{tab:permnull-app}
  \vspace{-3pt}
  \small
  \setlength{\tabcolsep}{4pt}
  \begin{tabular}{lccc}
    \toprule
    $T$ & observed $\rho$ & null $[2.5\%, 97.5\%]$ & one-sided $p$ \\
    \midrule
    $0.3$ & $+0.87$ & $[-0.49, +0.33]$ & $0.002$ \\
    $0.7$ & $+0.85$ & $[-0.51, +0.36]$ & $0.002$ \\
    $1.0$ & $+0.86$ & $[-0.53, +0.33]$ & $0.002$ \\
    \bottomrule
  \end{tabular}

\end{table}

\subsection{\revision{Feature recurrence}}
\label{app:exp-stability}

\paragraph{Full bootstrap-Lasso ranking.}
\revision{\Cref{tab:stability-full} gives selection frequencies for the 18 candidate features in the bootstrap-Lasso fits on 50 configurations. The 80\% line is a reporting convention. The compact ridge predictor instead uses four fixed inputs, including the position of the first correct sample. Computing that input requires gold answers regardless of how often Lasso selects it.}

\begin{table*}[h]
  \centering
  \caption{\revision{Bootstrap-Lasso recurrence across $500$ configuration resamples under grouped inner cross-validation. The $80\%$ marker is descriptive.}}
  \label{tab:stability-full}
  \vspace{-3pt}
  \small
  \setlength{\tabcolsep}{6pt}
  \begin{revised}
  \begin{tabular}{lcc}
    \toprule
    Feature & Selection freq.\ & Recurs $\geq80\%$? \\
    \midrule
    \texttt{per\_prompt\_answer\_entropy\_median} & $88.8\%$ & \checkmark \\
    \texttt{completion\_length\_variance} & $87.6\%$ & \checkmark \\
    \texttt{answer\_entropy} & $86.2\%$ & \checkmark \\
    \texttt{reasoning\_step\_count\_mean} & $84.2\%$ & \checkmark \\
    \texttt{majority\_size\_std} & $83.4\%$ & \checkmark \\
    \texttt{first\_correct\_sample\_position\_median} & $78.6\%$ & --- \\
    \texttt{self\_bleu} & $62.0\%$ & --- \\
    \texttt{mean\_topK\_mass} & $58.4\%$ & --- \\
    \texttt{completion\_repetition\_4gram} & $56.8\%$ & --- \\
    \texttt{embed\_sim} & $39.4\%$ & --- \\
    \texttt{seq\_lp\_spread} & $24.0\%$ & --- \\
    \texttt{majority\_size\_min} & $11.0\%$ & --- \\
    \texttt{mean\_logprob} & $1.8\%$ & --- \\
    \texttt{uniq\_2gram\_ratio} & $1.6\%$ & --- \\
    \texttt{majority\_fraction} & $1.4\%$ & --- \\
    \texttt{agreement\_rate} & $1.0\%$ & --- \\
    \texttt{chosen\_logprob\_disagreement\_pairwise} & $0.2\%$ & --- \\
    \texttt{std\_logprob} & $0.0\%$ & --- \\
    \bottomrule
  \end{tabular}
  \end{revised}

\end{table*}

\revision{Both entropy features, completion-length variance, mean reasoning-step count, and majority-fraction spread exceed 80\%. Median first-correct-sample position appears in 78.6\% of fits.}

\begin{figure}[t]
  \centering
  \includegraphics[width=\columnwidth]{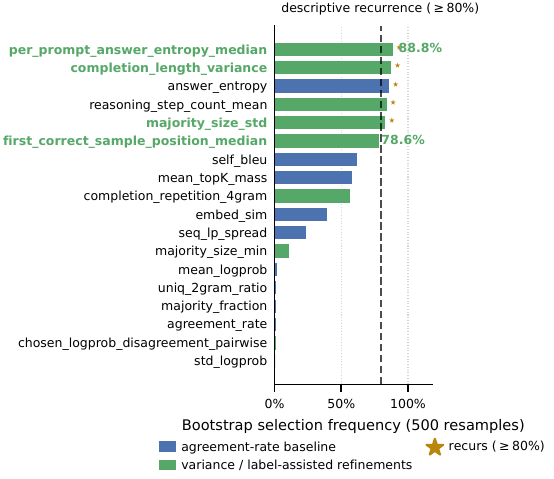}
  \vspace{-10pt}
  \caption{\revision{Bootstrap-Lasso selection frequencies with grouped inner cross-validation. The dashed line marks 80\%, and stars indicate features above it.}}
  \label{fig:stability}
  \vspace{-10pt}
\end{figure}

\subsection{Robustness}
\label{app:exp-robustness}

\paragraph{Across the scaling curve.}
\revision{For each candidate count $k'$, we recompute $\mathrm{BoN}@k'$ from the first $k'$ candidates and refit the 18-feature ridge on the earlier set of 32 configurations. The features and $\mathrm{pass}@1$ use the full pool of 64 candidates. \Cref{tab:k-sweep-app} reports the correlations by temperature from these joint fits.}

\begin{table}[h]
  \centering
  \caption{\revision{Candidate-count sweep for the 18-feature predictor on 32 configurations. Each row uses a joint ridge fit and reports LOSO Spearman correlations by temperature.}}
  \label{tab:k-sweep-app}
  \vspace{-3pt}
  \small
  \setlength{\tabcolsep}{4pt}
  \begin{tabular}{cccc}
    \toprule
    $k'$ & $T = 0.3$ & $T = 0.7$ & $T = 1.0$ \\
    \midrule
    2  & \revision{$0.64$} & \revision{$0.79$} & \revision{$0.67$} \\
    4  & \revision{$0.43$} & \revision{$0.77$} & \revision{$0.64$} \\
    8  & \revision{$0.56$} & \revision{$0.79$} & \revision{$0.78$} \\
    16 & \revision{$0.77$} & \revision{$0.87$} & \revision{$0.79$} \\
    32 & \revision{$0.82$} & \revision{$0.80$} & \revision{$0.76$} \\
    64 & $0.87$ & $0.85$ & $0.86$ \\
    \bottomrule
  \end{tabular}

\end{table}

\paragraph{Across gain functions.}
\revision{\Cref{tab:gain-robustness} fits the 18-feature ridge to five gain functions on the same 32 configurations.}

\begin{table}[h]
  \centering
  \caption{\revision{We fit the 18-feature joint ridge regression to each gain function on the same 32 configurations. LOSO holds out one checkpoint and task-domain group at a time. Subscripts are 95\% cluster-bootstrap half-widths. The normalized gain uses $\varepsilon_0=0.1$. Majority voting selects the most frequent extracted answer.}}
  \label{tab:gain-robustness}
  \vspace{-3pt}
  \small
  \setlength{\tabcolsep}{3pt}
  \begin{tabular}{lccc}
    \toprule
    Target $G$ & In $\mathcal{G}_L$? & Selector & LOSO $\rho$ \\
    \midrule
    $G_{\mathrm{add}}$ & yes & RM verifier & $\mathbf{+0.87_{\pm .15}}$ \\
    $G_{\mathrm{MV}}$ (true) & yes & plurality vote & $\phantom{+}0.00_{\pm .46}$ \\
    $G_{\mathrm{norm}}|_{\varepsilon_0}$ & yes (trunc.) & RM verifier & $\phantom{+}+0.88_{\pm .11}$ \\
    $G_{\mathrm{mult}}$ & no & RM verifier & $\phantom{+}+0.84_{\pm .19}$ \\
    $G_{\mathrm{log}}$ & no & RM verifier & $\phantom{+}+0.83_{\pm .20}$ \\
    \bottomrule
  \end{tabular}

\end{table}

\paragraph{\revision{Across post-training methods.}}
\revision{Holding out one post-training method at a time gives correlations from $0.78$ to $0.94$ (\cref{tab:per-method}).}

\begin{table}[h]
  \centering
  \caption{\revision{We fit the joint ridge regression on five post-training methods and evaluate on the held-out sixth. Subscripts are 95\% cluster-bootstrap half-widths.}}
  \label{tab:per-method}
  \vspace{-3pt}
  \small
  \setlength{\tabcolsep}{3pt}
  \begin{tabular}{@{}lcc@{}}
    \toprule
    \revision{Held-out method} & held-out $\rho$ & \revision{$n$ configs.} \\
    \midrule
    ORPO   & $\mathbf{0.94_{\pm .08}}$  & $11$ \\
    KTO    & $0.90_{\pm .18}$           & $8$  \\
    SimPO  & $0.89_{\pm .22}$           & $10$ \\
    DPO    & $0.88_{\pm .17}$           & $11$ \\
    SFT    & $0.85_{\pm .15}$           & $8$  \\
    GRPO   & $0.78_{\pm .32}$           & $8$  \\
    \bottomrule
  \end{tabular}

\end{table}

\revision{The cluster-bootstrap confidence intervals exclude zero for all six held-out methods. ORPO gives the highest correlation and the narrowest interval. GRPO gives the lowest correlation and the widest interval.}

\paragraph{Across reward models.}
\revision{We rescore the same samples with ArmoRM-Llama3-8B \citep{Wang2024ArmoRM} to measure gains for the same 56 configurations. In each split, the Skywork training folds determine the feature scaling, ridge penalty, and coefficients. We use that fit to predict gains for the held-out group and compare the predictions with the gains measured by ArmoRM (\cref{tab:cross-rm-target}).}

\paragraph{Sensitivity and ablation.}
\revision{\Cref{tab:sensitivity-ablation} checks the ridge penalty and standardization on the same 32 configurations. The correlation stays at $0.91$ for penalties from $10^{-3}$ to $10$ and is $0.88$ at $\alpha=100$. \Cref{tab:temperature-pooling} reports the temperature comparison on the same configurations.}

\begin{table}[h]
  \centering
  \caption{\revision{LOSO Spearman correlations for ridge variants on the same 32 configurations.}}
  \label{tab:sensitivity-ablation}
  \vspace{-3pt}
  \small
  \setlength{\tabcolsep}{4pt}
  \begin{tabular}{ll}
    \toprule
    Variant & LOSO $\rho$ \\
    \midrule
    Baseline ridge (4 features + $T$ covariate) & $0.91$ \\
    \midrule
    Ridge $\alpha = 10^{-3}$ (no CV) & $0.91$ \\
    Ridge $\alpha = 10^{-2}$ (no CV) & $0.91$ \\
    Ridge $\alpha = 10^{-1}$ (no CV) & $0.91$ \\
    Ridge $\alpha = 1$ (no CV) & $0.91$ \\
    Ridge $\alpha = 10$ (no CV) & $0.91$ \\
    Ridge $\alpha = 100$ (over-regularized) & $0.88$ \\
    \midrule
    Ridge without standardization & $0.91$ \\
    \bottomrule
  \end{tabular}

\end{table}

\subsection{Where the predictor fails}
\label{app:exp-failures}
\label{app:failure}

\paragraph{\revision{Leave-one-domain-out.}} \revision{The earlier 18-feature analysis includes math, programming, and reasoning configurations. Spearman correlation across all pooled held-out predictions is $+0.72$ (95\% bootstrap interval $[0.29,0.87]$). When the entire programming domain is held out, the correlation on programming configurations is $-0.56_{\pm .33}$.}

\paragraph{\revision{Large prediction errors.}}
\revision{\Cref{tab:adversarial-list} lists the five largest absolute LOSO residuals from the earlier 18-feature analysis, which included code. A residual is measured gain minus predicted gain. A positive residual indicates underprediction, and a negative residual indicates overprediction. The examples include both signs.}

\begin{table}[h]
  \centering
  \caption{\revision{Five largest absolute LOSO residuals in the exploratory 18-feature analysis. Each row is a configuration at one temperature.}}
  \label{tab:adversarial-list}
  \vspace{-3pt}
  \small
  \setlength{\tabcolsep}{4pt}
  \begin{tabular}{lc}
    \toprule
    Configuration & residual \\
    \midrule
    SFT-gemma-2-9b-code at $T = 1.0$           & $+0.11$ \\
    SFT-gemma-2-2b-math at $T = 1.0$           & $-0.09$ \\
    SFT-gemma-2-9b-code at $T = 0.7$           & $+0.09$ \\
    SFT-Llama-3.2-3B-math at $T = 1.0$         & $+0.07$ \\
    GRPO-Llama-3.1-8B-code at $T = 1.0$        & $-0.07$ \\
    \bottomrule
  \end{tabular}

\end{table}

\section{Cross-dataset BoN transfer}
\label{app:crossdata-bon}

\revision{We recompute the four features from 64 new samples per prompt on MATH500 \citep{Hendrycks2021Math, Lightman2023Lets} and HumanEval \citep{Chen2021Evaluating}, and measure gains with Skywork. We fit the four-feature ridge on GSM8K for the MATH500 test and on the HumanEval and MBPP pool for the HumanEval test. We apply each fit without retraining on the new generations. HumanEval reuses prompts from the original programming pool, so that test measures performance on fresh samples from overlapping prompts.}

\begin{table}[h]
  \centering
  \caption{\revision{Spearman correlations on MATH500 and HumanEval after recomputing features and measuring gains with Skywork.}}
  \label{tab:crossdata-bon}
  \vspace{-3pt}
  \small
  \setlength{\tabcolsep}{3pt}
  \begin{tabular}{lc}
    \toprule
    Benchmark & \revision{Spearman $\rho$} \\
    \midrule
    MATH500 & $+0.79$ ($p<10^{-4}$) \\
    HumanEval & $+0.35$ ($p=0.017$) \\
    \bottomrule
  \end{tabular}

\end{table}

\end{document}